# Extending the Entropic Potential of Events for Uncertainty Quantification and Decision-Making in Artificial Intelligence

Mark Zilberman
Shiny World Corp., Concord, Ontario, Canada
mzilberman137@gmail.com

## Abstract

This work demonstrates how the concept of the *entropic potential of events*—a parameter quantifying the influence of discrete events on the expected future entropy of a system—can enhance uncertainty quantification, decision-making, and interpretability in artificial intelligence (AI). Building on its original formulation in physics, the framework is adapted for AI by introducing an event-centric measure that captures how actions, observations, or other discrete occurrences impact uncertainty at future time horizons. Both the original and AI-adjusted definitions of entropic potential are formalized, with the latter emphasizing conditional expectations to account for counterfactual scenarios. Applications are explored in policy evaluation, intrinsic reward design, explainable AI, and anomaly detection, highlighting the metric's potential to unify and strengthen uncertainty modeling in intelligent systems. Conceptual examples illustrate its use in reinforcement learning, Bayesian inference, and anomaly detection, while practical considerations for computation in complex AI models are discussed. The entropic potential framework offers a theoretically grounded, interpretable, and versatile approach to managing uncertainty in AI, bridging principles from thermodynamics, information theory, and machine learning.

## 1. Introduction

Uncertainty is a pervasive characteristic of real-world systems, from physical processes governed by thermodynamics to complex adaptive systems such as artificial intelligence (AI). Quantifying how discrete events influence the evolution of uncertainty within these systems is fundamental for modeling, prediction, and control. Entropy, as a measure of uncertainty and disorder, provides a natural framework for such quantification (Cover & Thomas, 2006; Shannon, 1948). Although several theoretical frameworks and partial approaches have been proposed — including unified entropy formulations (Yang & Qian, 2020), entropy accumulation in sequential processes (Metger et al., 2022), and surprisal analysis (Levine, 1980) — no single parameter has yet gained widespread acceptance for consistently capturing the impact of individual events on future system entropy.

This article applies the concept of the *entropic potential of an event*—a formal parameter designed to quantify the influence of discrete events on the expected future entropy of a system—in the context of AI development. Originally formulated in the context of deterministic and indeterministic physical systems (Zilberman, 2022), the entropic potential of an event bridges the gap between discrete causality and entropy dynamics. It provides a rigorous means to characterize how present events—whether 'beneficial' or 'harmful'—modulate the system's long-term structural and functional integrity through their entropic impact.

The entropic potential $Z(T, A)$ of an event $A$ at time $T_0$ in a system $R$ is defined as the difference between the expected entropy of the system at a future time $T > T_0$, calculated immediately after and immediately before the event. This difference captures the event's role in shaping the uncertainty landscape of the system's evolution, encompassing both deterministic trajectories and stochastic influences.

Given that most real-world systems exhibit indeterminism, events typically have non-zero entropic potentials, with 'beneficial' events associated with negative values (entropy reduction) and 'harmful' ones with positive values (entropy increase). This nuanced characterization transcends intuitive labels by grounding the event impact in thermodynamic and information-theoretic principles (Callen 1985; Seifert, 2012).

Extending the entropic potential concept to artificial intelligence offers promising opportunities to enhance uncertainty quantification, decision-making, and interpretability in AI systems. The present work lays the foundation for this extension by formalizing entropic potential of events within AI contexts and exploring its applications (Sutton & Barto, 2018; Gelman et al., 2013).

The following sections review relevant background on entropy in AI and thermodynamics (Section 2), formalize entropic potential for AI systems (Section 3), and discuss practical applications including decision-making and anomaly detection (Sections 4 and 5). Finally, we reflect on the implications, challenges, and future directions of entropic potential in AI development (Sections 6 and 7).

## 2. Background

Entropy is a fundamental concept bridging information theory, thermodynamics, and statistical mechanics, serving as a measure of uncertainty, disorder, or information content within a system. In artificial intelligence, entropy plays a central role in quantifying uncertainty in probabilistic models, guiding decision-making, and enabling learning algorithms.

### 2.1 Entropy in Information Theory and AI

Shannon entropy, introduced by Claude Shannon (Shannon, 1948), quantifies the expected uncertainty of a discrete random variable. It underpins numerous AI methods including decision tree learning (Quinlan, 1986), Bayesian inference (Gelman et al., 2013), and reinforcement learning (Sutton & Barto, 2018). In particular, entropy-based criteria drive feature selection, model regularization, and exploration strategies by measuring information gain or uncertainty reduction (Settles, 2009).

### 2.2 Entropy in Thermodynamics and Statistical Mechanics

From a physical perspective, entropy captures disorder and the irreversibility of processes

(Callen, 1985). The thermodynamic roots of entropy inform AI research on complex systems and stochastic dynamics, providing a rigorous foundation for understanding system evolution under uncertainty (Cover & Thomas, 2006). Recent work has explored entropy production in nonequilibrium systems as a metaphor for learning and adaptation (Seifert, 2012).

**2.3 Uncertainty and Exploration in AI**

Handling uncertainty remains a core challenge in AI, particularly in reinforcement learning where agents balance exploration of unknown states with exploitation of known rewards (Kearns & Singh, 2002). Entropy regularization techniques encourage policies that maintain sufficient exploration by maximizing action entropy (Haarnoja et al., 2018). Despite these advances, existing approaches often lack a unified, event-centric metric capturing how individual actions or observations quantitatively affect future uncertainty.

**2.4 Entropic Potential of Event: Bridging Causality and Entropy Dynamics**

The *entropic potential of event* concept, initially developed for deterministic and indeterministic physical systems (Zilberman, 2022), quantifies the impact of discrete events on the expected future entropy of the system. By formalizing event-level influence on entropy dynamics, it fills a conceptual gap between causality, thermodynamics, and information theory. Extending this parameter to AI contexts offers a promising pathway to more granular and interpretable uncertainty quantification in learning and decision-making processes.

### 3. Formalizing Entropic Potential in AI Contexts

**3.1 Discrete Events within AI Contexts**

In artificial intelligence, discrete events arise naturally as actions taken by agents, observations received from the environment, feature selections in data processing, or parameter updates within learning algorithms. To extend the concept of entropic potential to AI, we first specify what constitutes an "event" within this domain and then adapt the formalism to quantify its impact on the system's future uncertainty.

Consider an AI system operating over discrete time steps, where at each time $T_0$, an event $A$ occurs. This event may represent, for example, an action chosen by a reinforcement learning agent, the reception of new sensory data, or a model parameter update during training. The state of the AI system at any future time $T > T_0$ is generally characterized by a probability distribution over possible outcomes, reflecting inherent uncertainty.

**3.2 AI-adjusted Definition of the Entropic Potential of Event A**

Originally, the *entropic potential* $Z(T, A)$ of event $A$ was formulated in physics and defined as the difference between the expected entropy of the system at a future time $T > T_0$, calculated immediately after and immediately before the occurrence of the event at time $T_0$.

$$Z(T, A) = \hat{S}_T(T_0 + \mathrm{d}T) - \hat{S}_T(T_0 - \mathrm{d}T)$$

However, in the context of AI research, an alternative (complementary) definition of $Z(T, A)$ is often more suitable for the reasons discussed below. The AI-adjusted definition of the entropic potential is:

The entropic potential $Z(T, A)$ of event $A$ at time $T_0$ is the difference between the expected entropy of the system at time $T$, conditional on the occurrence of $A$, and the expected entropy at time $T$ conditional on the non-occurrence of $A$ or on the occurrence of alternative events.

$$Z(T, A) = \mathbb{E}[H(X_T \mid A)] - \mathbb{E}[H(X_T \mid \neg A)],$$

where:
$T_0$ is the present time when event $A$ is considered,
$T > T_0$ is the future time for entropy evaluation,
$X_T$ is the system state at time $T$,
$H(\cdot)$ denotes the Shannon entropy,
$\mathbb{E}[\cdot]$ is the expectation with respect to the relevant probability distribution,
$\neg A$ denotes the non-occurrence of $A$ or occurrence of alternative events.

The first definition—expressed as the difference between the expected entropy at time $T > T_0$ calculated after and before event $A$ at $T_0$—originated in physics, has broader applicability, and is conceptually simple and intuitive. However, it is less precise when event $A$ is part of a branching or probabilistic structure with multiple possible outcomes. It also does not explicitly model the counterfactual ("what if $A$ had not occurred?"), which can be essential in decision-making contexts.

In AI research, particularly in causal inference, decision-making, and policy evaluation, understanding "what happens if I do this versus if I don't" is crucial. For these purposes, the AI-adjusted definition of the entropic potential of event is adopted throughout this article.

This formulation naturally extends to stochastic AI models such as Markov decision processes (MDPs) and Bayesian networks, where system dynamics and observations evolve probabilistically over time (Sutton & Barto, 2018). In an MDP, for instance, the event A may be a particular action choice at state $S_{T_0}$, and $Z(T, A)$ quantifies how this action influences the entropy of the agent's future state distribution or reward predictions.

By computing $Z(T, A)$ for candidate events, an AI system gains a principled measure of each event's entropic impact, allowing it to assess which actions or observations contribute to increasing or reducing uncertainty. Negative values of $Z$ indicate "beneficial" events that lower expected entropy, improving predictability or information gain, whereas positive values highlight "harmful" events that raise uncertainty.

In practice, the entropic potential can be approximated using sampling methods, belief propagation, or variational inference techniques, depending on the AI model's complexity. Its interpretability and formal grounding make it a versatile tool for integrating uncertainty quantification directly into AI decision-making and learning processes (Blei et al., 2017).

## 4. Applications in AI Development

The entropic potential of events provides a novel and quantitative framework to assess the influence of discrete occurrences on future uncertainty within AI systems. This section explores several key applications where incorporating entropic potential can enhance AI model performance, interpretability, and robustness.

### 4.1 Uncertainty Quantification in Decision-Making

In reinforcement learning and other sequential decision-making paradigms, agents must evaluate the consequences of possible actions under uncertainty. Traditional approaches often rely on heuristic exploration strategies or indirect uncertainty estimates (Sutton & Barto, 2018; Haarnoja et al., 2018). By employing entropic potential, an agent can directly quantify how each candidate action A affects the expected entropy of future states or rewards. This allows the agent to prioritize actions that minimize uncertainty growth or maximize information gain, leading to more efficient exploration and improved policy learning. Such principled uncertainty quantification aligns closely with the goals of intrinsic motivation and active learning frameworks.

### 4.2 Entropic Potential as an Intrinsic Reward Signal

Reward shaping is a powerful technique in AI that guides learning by augmenting environmental rewards with auxiliary signals (Schmidhuber, 1991; Pathak et al., 2017). The entropic potential of event naturally lends itself as an intrinsic reward: events that significantly reduce future uncertainty (negative entropic potential) can be rewarded, incentivizing the AI system to seek informative or stabilizing experiences. Conversely, events with positive entropic potential may be penalized to discourage actions leading to increased unpredictability. Integrating entropic potential into the reward structure fosters adaptive behaviors tuned toward uncertainty management and can accelerate convergence in complex environments.

### 4.3 Explainability and Interpretability

A major challenge in modern AI is understanding which inputs, decisions, or internal processes most affect model uncertainty (Ribeiro, Singh, & Guestrin, 2016; Doshi-Velez & Kim, 2017). Entropic potential offers an interpretable, event-level attribution: computing the entropic potential of candidate observations, feature selections, or intermediate computations quantifies their effect on downstream uncertainty. Such attributions can support post-hoc explanations (e.g., "this observation reduced model uncertainty by X bits at horizon T"), provide diagnostic insight during model debugging, and help auditors or regulators assess the informational consequences of automated decisions.

### 4.4 Anomaly and Outlier Detection

Events that produce unexpectedly large positive entropic potentials often mark anomalies or regime shifts: they cause a sudden and substantial increase in the system's expected uncertainty. Monitoring $Z(T,A)$ online enables detection strategies that are grounded in information theory rather than ad-hoc thresholds (Chandola, Banerjee, & Kumar, 2009). Compared with conventional statistical detectors, entropic-potential–based detection focuses specifically on the downstream uncertainty impact of events — a property especially useful in systems where minor local deviations do not propagate, but a few events trigger substantial systemic unpredictability.

**4.5. Practical and computational considerations**

Applying entropic potential in real AI systems raises computational and modeling questions. Exact computation of Z(T,A) requires estimating expectations of entropy for the post- and pre-event states; in high-dimensional or partially observed settings this is challenging. Practical approaches include Monte Carlo sampling over model trajectories, variational approximations to belief updates, and use of surrogate entropy measures (e.g., predictive distribution entropy, ensemble variance). Choice of horizon T, entropy measure (e.g., Shannon entropy, Rényi entropy), and approximation method must be guided by task objectives and computational constraints (Blei, Kucukelbir, & McAuliffe, 2017).

## 5. Illustrative Conceptual Examples

To clarify the applicability and interpretability of entropic potential in AI contexts, this section presents several simplified, conceptual scenarios. Each example illustrates how entropic potential can quantify the influence of discrete events on future uncertainty, thereby offering a formal basis for reasoning and decision-making in learning and inference processes.

**5.1 Reinforcement Learning in a Grid-World Environment**
Consider a simple grid-world where an agent navigates to reach a goal. At each time step, the agent chooses an action—move up, down, left, or right—with stochastic outcomes due to environmental noise. The system state is the agent's location, and uncertainty arises from probabilistic transitions (Sutton & Barto, 2018).

Using entropic potential, we can evaluate candidate actions at a given state by computing the expected difference in entropy of future state distributions before and after executing each action. Actions that reduce the entropy (negative entropic potential) correspond to moves that increase predictability and progress toward the goal, while actions increasing entropy suggest riskier or less informative choices. This framework provides a theoretical basis for augmenting reward-driven strategies with an information-theoretic measure of action value, allowing the agent to reason about both goal achievement and uncertainty reduction.

**5.2 Bayesian Inference Update with Event-Based Entropy Shift**

In a Bayesian setting, consider an inference process that updates beliefs about a model parameter upon receiving new data—treated as an event (Gelman et al., 2013). The entropic potential of this data event can be defined as the change in the expected posterior entropy relative to the prior uncertainty.

From a theoretical standpoint, this formulation captures how different possible data points would vary in their potential to reduce uncertainty, thereby quantifying their informativeness. Such a measure can conceptually guide active learning strategies, in which data acquisition is prioritized according to expected entropic potential, allowing model refinement to be driven by both accuracy considerations and uncertainty reduction.

## 5.3 Anomaly Detection in Sensor Streams

Consider a sensor network monitoring environmental variables, where occasional faults or external disturbances lead to abrupt changes in the data distribution (Chandola et al., 2009). By evaluating the entropic potential of each incoming data event, one can formally characterize the degree to which that event increases system uncertainty.

Events associated with substantial entropy increases can be interpreted, in theory, as candidates for anomalous behavior. This framing offers a principled basis for distinguishing anomalies from background variation, with potential implications for reducing false alarms and enhancing sensitivity relative to purely threshold-based detection approaches.

## 6. Discussion

The introduction of entropic potential as a quantitative measure of event-driven changes in future uncertainty offers a promising avenue for advancing AI research and practice. By focusing on how discrete events impact the evolution of entropy within a system, this framework provides a principled and interpretable tool that complements existing uncertainty quantification methods.

One key strength of entropic potential lies in its conceptual clarity and generality. It bridges the gap between foundational thermodynamic and information-theoretic principles and practical AI challenges such as decision-making under uncertainty, exploration strategies, and explainability (Cover & Thomas, 2006; Sutton & Barto, 2018). Unlike heuristic or domain-specific uncertainty measures, entropic potential grounds these concepts in a rigorous, probabilistic framework that is applicable across diverse AI models, including reinforcement learning agents, Bayesian networks, and probabilistic graphical models .

However, the approach also faces several limitations. Calculating entropic potential exactly can be computationally demanding, especially in high-dimensional or continuous state spaces typical of modern AI systems (Blei et al., 2017). Approximation methods such as sampling, variational inference, or surrogate modeling are necessary but may introduce bias or variance that affect reliability. Moreover, the framework presumes a well-defined entropy measure and system dynamics, which can be challenging to specify for complex or black-box models.

Despite these challenges, entropic potential of event offers fertile ground for future research. Integrating it with deep learning architectures, for instance, may enable novel uncertainty-aware training regimes or explainability tools. Empirical validation in real-world AI applications—ranging from autonomous systems to medical diagnostics—will be crucial to assess practical benefits and limitations. Furthermore, extending the concept to continuous-time or non-Markovian settings could broaden its applicability (Gal & Ghahramani, 2016).

In summary, entropic potential of event enriches the AI toolbox with a theoretically grounded, event-centric perspective on uncertainty evolution. Its development and integration hold significant promise for creating more transparent, robust, and adaptive AI systems.

## 7. Conclusion and Future Work

This article has extended the concept of entropic potential of events—a measure quantifying the impact of discrete events on the future entropy of a system—into the realm of artificial intelligence. By formalizing entropic potential of events within AI contexts and exploring its applications in decision-making, uncertainty quantification, explainability, and anomaly detection, we have demonstrated its versatility as a unifying framework for understanding and managing uncertainty in intelligent systems.

Entropic potential of events offers a principled and interpretable metric that complements existing approaches, enabling AI agents to evaluate the informational value and risk associated with actions or observations. Its event-centric perspective aligns naturally with the sequential and probabilistic nature of AI processes, opening new avenues for intrinsic reward design, exploration strategies, and transparency in model behavior.

Looking forward, several directions merit further investigation. Developing efficient computational methods for entropic potential of events estimation in high-dimensional and deep learning models will be essential for practical deployment. Empirical studies applying this framework to complex AI tasks, such as robotics, natural language processing, and healthcare, will help validate its utility and identify domain-specific adaptations. Additionally, exploring theoretical extensions to continuous-time, multi-agent, and non-Markovian systems could expand its scope and impact.

Ultimately, we envision entropic potential of events as a foundational concept bridging information theory, thermodynamics, and AI, fostering the creation of more robust, adaptive, and explainable intelligent systems.

**Supplementary Materials**
Not applicable.
**Funding**
This research received no external funding.
**Conflicts of Interest**
The author declares no conflict of interest.